# Topological Feature Search Method for Multichannel EEG: Application in ADHD classification

**Tianming Cai[1], Guoying Zhao[1, 2], Junbin Zang[1, 2, *], Chen Zong[3], Zhidong Zhang[1, *], Chenyang Xue[1]**

[1] North University of China, School of Instrument and Electronics, No.3 College Road, Jiancaoping District, Taiyuan, Shanxi, 030051, China

[2] Shanxi College of Technology, No.11 Changning Street, Development Zone, Shuozhou, Shanxi, 036000, China

[3] The Second Hospital of Shanxi Medical University, No.382 Wuyi Road, Taiyuan, Shanxi, 030001, China

**Abstract：**

In recent years, the preliminary diagnosis of Attention Deficit Hyperactivity Disorder (ADHD) using electroencephalography (EEG) has attracted the attention from researchers. EEG, known for its expediency and efficiency, plays a pivotal role in the diagnosis and treatment of ADHD. However, the non-stationarity of EEG signals and inter-subject variability pose challenges to the diagnostic and classification processes. Topological Data Analysis (TDA) offers a novel perspective for ADHD classification, diverging from traditional time-frequency domain features. However, conventional TDA models are restricted to single-channel time series and are susceptible to noise, leading to the loss of topological features in persistence diagrams.This paper presents an enhanced TDA approach applicable to multi-channel EEG in ADHD. Initially, optimal input parameters for multi-channel EEG are determined. Subsequently, each channel's EEG undergoes phase space reconstruction (PSR) followed by the utilization of k-Power Distance to Measure (k-PDTM) for approximating ideal point clouds. Then, multi-dimensional time series are re-embedded, and TDA is applied to obtain topological feature information. Gaussian function-based Multivariate Kernel Density Estimation (MKDE) is employed in the merger persistence diagram to filter out desired topological feature mappings. Finally, the persistence image (PI) method is employed to extract topological features, and the influence of various weighting functions on the results is discussed.The effectiveness of our method is evaluated using the IEEE ADHD dataset. Results demonstrate that the accuracy, sensitivity, and specificity reach 78.27%, 80.62%, and 75.63%, respectively. Compared to traditional TDA methods, our method was effectively improved and outperforms typical nonlinear descriptors. These findings indicate that our method exhibits higher precision and robustness.

## 1. Introduction

Attention Deficit Hyperactivity Disorder is a severe neurological condition primarily affecting children and adolescents, with a lesser incidence in adults [1]. Symptoms of ADHD typically manifest as inattention, impulsivity, irritability, and restlessness. Investigations indicate that individuals with ADHD often present comorbidities such as anxiety disorders, conduct disorders, and Oppositional Defiant Disorder (ODD) [2]. The high prevalence of comorbidities complicates the management of ADHD-related treatments, possibly contributing to the significant increase in non-natural mortality rates associated with ADHD in recent years [3].

As a convenient and rapid technological approach, non-invasive electroencephalography (EEG) finds widespread application in the diagnosis of various neurological disorders such as epilepsy [4], depression [5], and Alzheimer's disease [6]. Compared to other investigative modalities like fMRI, fNIRS, or EOG, EEG offers brain activity-related information that aids researchers in promptly identifying abnormal patterns in patient brains, a crucial aspect in the diagnosis of neurological disorders. Naturally, EEG has been introduced into screening and diagnosing ADHD [7]. Early research on using EEG for ADHD can be traced back to J. Lubar, who first observed increased θ activity accompanied by decreased β power in ADHD patients [8].

Analysing EEG data in both the time and frequency domains is a common approach. Typically, researchers extract relevant feature information from a set of EEG time series signals in the time-frequency domain and employ machine learning classifiers for classification. For example, Danlei Gu et al. [9] proposed Cross-Frequency Symbol Convergence Analysis (CEEMDAN CF-SCCM) based on Complete Ensemble Empirical Mode Decomposition with Adaptive Noise (CEEMDAN) to discern phase-amplitude coupling differences in various brain regions of ADHD patients. Anika Alim et al. [10] extracted Hjorth parameters, signal skewness, kurtosis, and entropy as features for ADHD signals. Joy C et al. [11] utilized the tunable Q-factor wavelet transform to extract frequency domain features of EEG in different frequency bands and effectively classified them using an ANN classifier.

On the other hand, researchers have also observed changes in brain functional connectivity in ADHD patients, making connectivity analysis of brain networks a crucial area of study. Abbas et al. [12] employed transfer entropy as a measure of information transmission to detect pairwise directional information transfer between EEG signals. H Kiiski et al. [13] used the Weighted Phase Lag Index to compute functional connectivity among all scalp channels. Cura et al. [14] employed the Intrinsic Time Decomposition (ITD) method to analyze EEG activity in ADHD children and extracted various connectivity features.

Furthermore, deep learning and its variations have found extensive applications in ADHD classification. Moghaddari M et al. [15] extracted samples of different frequency bands from EEG and formed two-dimensional images, subsequently employing a 13-layer two-dimensional CNN model for classification. Chang Y et al. [16] utilized Long Short-Term Memory (LSTM) networks based on EEG to learn cognitive state transitions and differentiate between ADHD and Neurotypical (NT) individuals.

Due to the ultra-high-dimensional nature of EEG signals, they exhibit nonlinear dynamics, which limits the effectiveness of linear techniques in signal detection [17]. Consequently, various nonlinear features of the EEG have become another focal point of research. Common nonlinear features such as fuzzy entropy (FE) [18], Lyapunov exponents (LE) [19], and fractal dimension (D2) [20], among others, have been applied in ADHD classification. In addition, it has been found that combining nonlinear analysis methods with brain functional connectivity represents a promising area of research. Ansarinasab [21] was the first to use the nonlinear Correlation between Probability of Recurrences (CRP) method to reveal changes in the brain network topology of ADHD patients. In addition, Ansarinasab studied the differences in the synchronization of simulated functional brain networks of ADHD patients under different emotional states based on the Hindmarsh-Rose neuron model [22]. Kaur [23] used correlation-based feature selection and particle swarm optimization to screen the Euclidean distance features in the EEG reconstructed phase space and evaluated the system using five machine learning methods. Across various tests, nonlinear descriptors consistently reveal the non-stationary and chaotic behavior of acquired brain signals.

Recently, there has been a surge in research using Topological Data Analysis (TDA) as a novel technique to represent the geometric structure of point clouds, offering new insights into extracting nonlinear information from EEG signals [24]. By leveraging persistent homology tools, TDA can unearth hidden topological features within signal point clouds and quantitatively represent them through persistence diagrams and their derived markers. Thus far, TDA has started to be applied in EEG analysis, including studies on neurodegenerative diseases [25], brain state recognition [26], as well as emotion recognition and classification [24]. However, these studies have yet to address the limitations of TDA. TDA is susceptible to noise interference [27], where signal noise mapped onto point clouds can lead to loss of topological features and misidentification. Furthermore, TDA is only applicable to single time series, which may reduce efficiency and accuracy in EEG analysis with multiple time series.

In summary, this paper aims to accurately identify hidden topological features in multi-channel EEG time series of ADHD patients through an improved TDA method. These features will serve as compelling evidence to distinguish ADHD patients from normal individuals, thereby enhancing early screening and diagnosis of ADHD. Initially, optimal parameters for multi-channel phase space reconstruction are determined, followed by the reconstruction of ideal point clouds based on k-Power Distance to Measure (k-PDTM) in single-channel mappings. The point clouds are then remapped and the entire phase space is subjected to a topological nonlinear analysis for feature extraction. The segmented features are then mapped to persistence diagrams (PD), and Gaussian function-based Multivariate Kernel Density Estimation (MKDE) filtering is applied to obtain the final PD. Finally, the Persistence Image (PI) method with different weighting functions is employed to further extract topological features. The specific framework of the proposed method is illustrated in Figure 1. The main contributions of this paper are as follows:

**1).** We employed nonlinear topological data analysis methods to elucidate the multichannel EEG alterations in patients with ADHD and validated our findings using the IEEE ADHD dataset.

2). We integrated k-PDTM and MKDE techniques to enhance the original TDA models, addressing the issue of topological feature loss in the multichannel EEG data of ADHD patients. Additionally, we devised a weighting function within the PI method to further extract topological features.

The main contents of this paper are as follows: Section 2 provides a detailed explanation of the dataset used and the corresponding preprocessing methods. Section 3 elaborates on our proposed method. Section 4 showcases the results of the experiments and relevant discussions. Finally, Section 5 concludes the paper.

## 2. Material and Signal Preprocessing

In this section, we will briefly introduce the basic information of the data we used, including its source and collection methods. Subsequently, we will preprocess the raw data to ensure its usability. Finally, we will determine necessary parameters in preparation for explaining the method in Section 3.

### *2.1. Data Materials*

The IEEE ADHD database used in this experiment is available on IEEE DataPort [28]. This database comprises 61 children diagnosed with ADHD and 60 healthy control subjects, consisting of boys and girls aged 7-12 years. The ADHD sample consisted of 48 boys and 13 girls with a mean age of 9.62 ± 1.75 years old; the healthy sample consisted of 50 boys and 10 girls with a mean age of 9.85 ± 1.77 years old. ADHD-diagnosed children were diagnosed by experienced psychiatrists according to DSM-IV criteria and were all administered methylphenidate for a duration of 6 months. Control group children were mentally normal, with no history of psychiatric disorders, epilepsy, or reports of any high-risk behaviors.

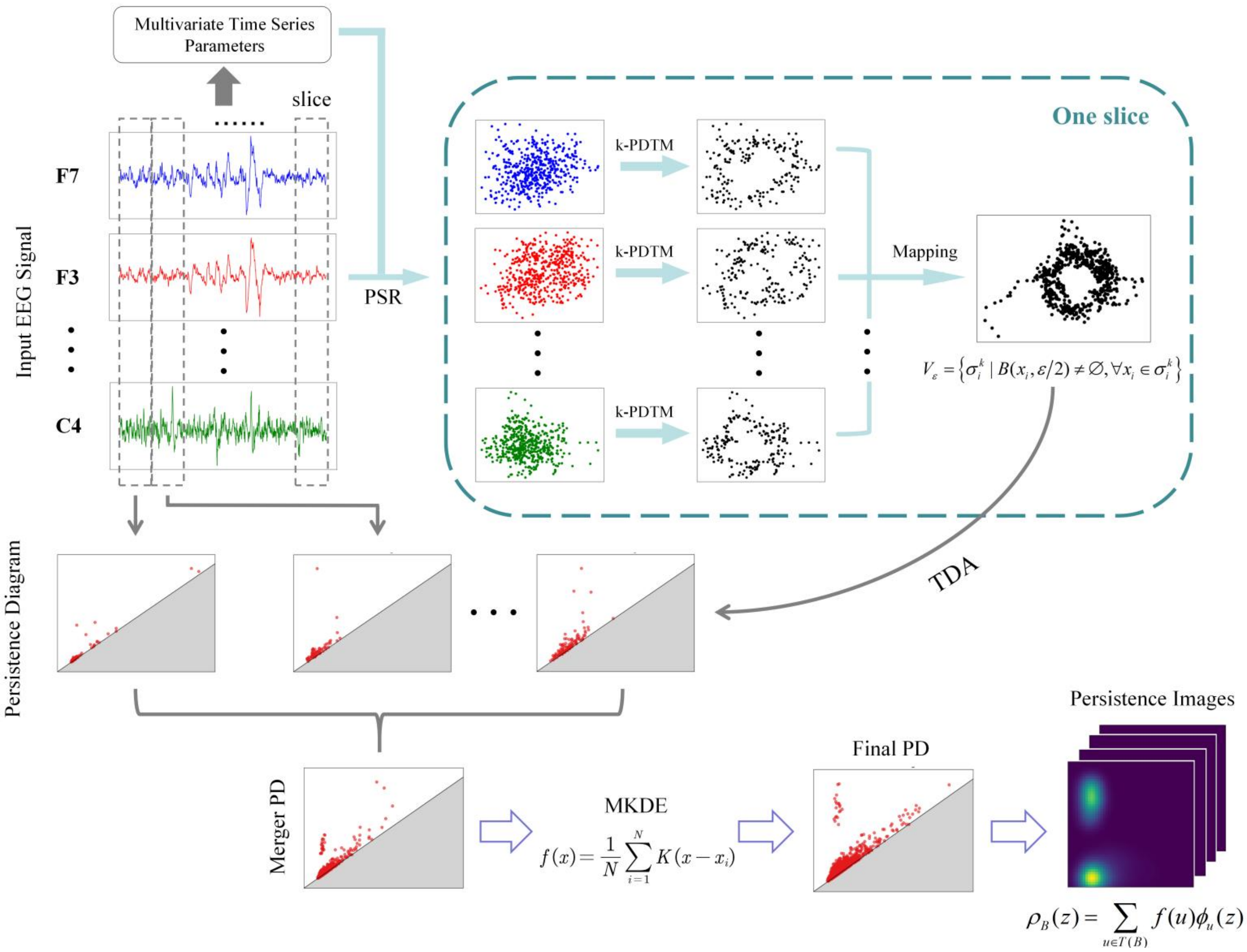

**Figure 1.** Framework of the proposed methodology

The EEG signals were recorded at the Roozbeh Hospital Center for Psychology and Psychiatry Research in Tehran, Iran. The electrodes were placed according to the international 10-20 system, totaling 19 EEG electrodes (Fz, Cz, Pz, C3, T3, C4, T4, Fp1, Fp2, F3, F4, F7, F8, P3, P4, T5, T6, O1, O2), with a sampling frequency of 128 Hz, The A1 and A2 electrodes are reference electrodes positioned on the earlobes. The EEG recording protocol was based on a visual attention task. In the task, children were presented with a series of cartoon images and asked to count the number of characters. The number of characters in each image was randomly selected between 5 and 16, and the size of the images was large enough for children to easily see and count. To ensure continuous stimuli during signal recording, each picture was displayed without interruption immediately after the child's response. Therefore, the duration of EEG recording throughout the cognitive visual task depended on the child's response speed. The minimum task duration was 50s for one of the control group and the maximum task duration was 285s for one subject with ADHD.

### *2.2. Data Preprocessing*

In the original signals, there are varying levels of noise and artifacts, thus necessitating denoising and artifact removal from the signals in the database. Since the earlobes reference electrodes (A1, A2) were already used for average referencing during data collection, and the 10-20 acquisition system is not a high-density recording system, the uneven electrode distribution is not suitable for average referencing of all scalp electrodes. Therefore, we did not perform a re-referencing process during data handling.

We employed a 4th-order Butterworth bandpass filter with cutoff frequencies from 0.5 Hz to 50 Hz for bandpass filtering and utilized Independent Component Analysis (ICA) to remove ocular artifacts from the signals [29]. Additionally, to reduce computational complexity and enhance efficiency, we did not utilize all EEG channels in subsequent experiments but rather focused on relevant EEG channels that play a significant role in ADHD. By conducting independent t-tests to calculate the average p-values of channel features, Maniruzzaman et al. identified 6 relevant electrodes [30]: Fz, F8, F3, C4, C3, and F7.

In the experiment, we segmented each channel of EEG into segments of 4 seconds in length, with each segment containing 512 sampling points. The selection of 4-second EEG segments is based on balancing the capture of sufficient information and maintaining a manageable data size for analysis. During the subsequent phase space reconstruction, the number of sampling points also affects the mapped phase space topology. Fewer sampling points cannot fully unfold the phase space structure (e.g., a 3-second segment contains only 384 sampling points), while too many sampling points can lead to information loss in the phase space structure (e.g., a 5-second segment contains up to 640 sampling points). In addition, due to variations in the experimental duration for each subject, the total number of segments differed for each participant.

It is worth noting that we chose to analyze the entire frequency band rather than individual bands. Although individual bands have been shown to be associated with ADHD in the frequency domain, analyzing the entire frequency band provides a comprehensive view of brain region activity in nonlinear analysis. Different frequency bands in the brain may interact on the same time scale, and using the entire frequency band allows us to capture these complex interactions and global dynamical features. Additionally, separating frequency bands might lead to information loss or overlook interactions between certain bands. Using the entire frequency band maximizes the retention of information from the original signal.

### *2.3. Determining the Parameters of Phase Space Reconstruction*

Phase space reconstruction (PSR) is a crucial preprocessing step in topological data analysis, as it transforms time series signals into point cloud structures, with time delay embedding being a necessary process to obtain the signal's point cloud [31]. In time delay embedding, the most critical parameters are the embedding dimension $m$ and time delay $\Delta t$ . In univariate time series, various methods exist to find the optimal embedding parameters, such as False Nearest Neighbors (FNN) [32], autocorrelation coefficients [33], and Mutual Information (MI) [34]. However, multivariate time series contain more information, thus, whether these methods can be extended to multivariate cases to recover the dynamical system of the time series is a question worth discussing [35].

When extending single-variable time series to multivariate conditions, the available time series consist of n-dimensional variables $\{x_i(t) \mid i = 1, 2, ..., n\}$ . Assuming $m_i$ and $\tau_i$ are the embedding dimension and delay time, respectively, for the $i-$th time series, the embedding vector for the $i-$th time series is given by:

$$X_i(t) = [x_i(t), x_i(t-\tau_i), ..., x_i(t-(m_i-1)\tau_i)] \tag{1}$$

There are various methods to determine parameters for multivariate time series, such as non-uniform state space reconstruction [36], local constant methods [37], or multivariate C-C methods [38]. These methods offer the advantage of avoiding irrelevant and redundant variables, but they come with a high computational burden and may disrupt some phase space features of single-variable time series, which is unfavorable for EEG multivariate signals. In nonlinear analysis, we aim to preserve the attractor structures of different single-variable time series from various brain regions. Garcia [39] proposed that in multivariate phase space reconstruction, methods based on single-variable analysis can be extended, with the resulting parameters seen as a balance of the parameters for each single variable. These methods can reconstruct attractors with similar topology to those of single-variable sequences. Therefore, we opt for the uniform multivariate average mutual information method [40] and the multidimensional extension method of FNN [41] to calculate the time delay and embedding dimension of multidimensional EEG signals. The optimal parameters calculated for each time series in multidimensional embedding are $m = 2$ and $t = 10$.

## 3. Methodology

After PSR, ADHD time series signals are transformed into point cloud form. Conducting Topological Data Analysis on this point cloud can effectively extract the nonlinear topological features of the original signal, with Persistent Homology (PH) being the core of TDA methods. In this section, we will briefly introduce the principles of topological data analysis and analyze why the direct application of TDA is not suitable for searching the topological features of ADHD. Finally, appropriate improvements are proposed.

### *3.1. Persistent Homology and Persistence Diagram*

Persistent homology feature extraction in a given point cloud relies on the simplicial complex $\mathcal{X}(\mathcal{V},\mathcal{S})$, which consists of a set of vertices $\mathcal{V}$ and a set of subsets of vertices $\mathcal{S}$ [42]. A k-simplex serves as the foundation of the simplicial complex, where each k-simplex contains k+1 points and can be abstractly represented as $\sigma_i^k = \{v_{i0}, v_{i1}, ..., v_{ik}\}$. The face of a k-simplex is the boundary of the simplex, which behaves as a subset of the (k-1) simplex, denoted as $\{v_{i0}, ..., v_{i_{j-1}}, v_{i_{j+1}}, ..., v_{ik}\}$. where $0 \le j \le k$. Therefore, each k-simplex also possesses k+1 faces.

As shown in Figure 2(a), a 0-simplex consists of one point and one face (itself), a 1-simplex consists of two points and two faces (the boundary of a 1-simplex is represented in point form, which is a subset of the 0-simplex), a 2-simplex consists of three points and three faces (the boundary of a 2-simplex is in line form, a subset of the 1-simplex), and a 3-simplex consists of four points and four faces (the boundary of a 3-simplex is in surface form, a subset of the 2-simplex), and so on.

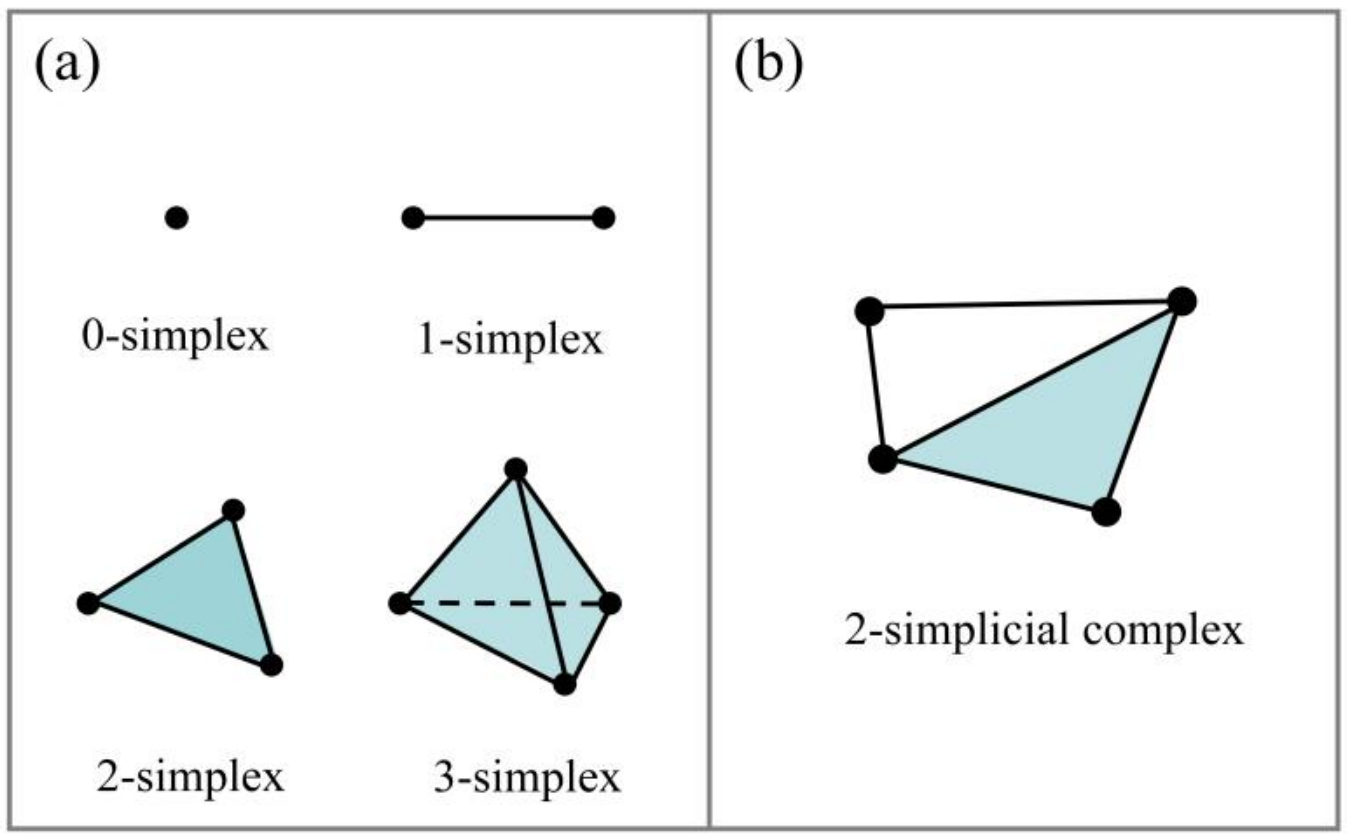

**Figure 2.** The intuitive geometric structures of simplex and simplicial complex. (a). The simplex of dimensions 0 to 4 and their inclusion relationships. (b). 2-simplicial complex comprised of four 0-simplex, five 1-simplex, and one 2-simplex.

The definition of a simplicial complex $\mathcal{X}$ is a finite set of simplex, which must satisfy two conditions:

- any face of $\sigma_i^k \in \mathcal{X}$ is also in $\mathcal{X}$;
- if $\sigma_1^k, \sigma_2^q \in \mathcal{X}$, then $\sigma_1^k \cap \sigma_2^q$ is a face for both $\sigma_1^k$ and $\sigma_2^q$.

The dimension of a simplicial complex is the maximum dimension of any simplex within it. Every simplex of order k can be associated with a geometric simplicial complex, embedded in a space of dimension at least k, such that their topological properties are equivalent [43]. As illustrated in Figure 2(b), 2-simplicial complex consists of simplex of order less than or equal to 2, embedded in a 2-dimensional space.

Constructing a simplicial complex from a given point cloud requires the introduction of a distance parameter. Let $X$ be a point cloud embedded in an $m$ dimensional space and $x_i (i = 1, 2, ..., n) \in X$. We define a ball $B$ centered at $x_i$ with radius $\varepsilon$. The Vietoris-Rips complex, denoted as $V_\varepsilon$, is a commonly used simplicial complex structure in TDA, satisfying the following definition:

$$V_\varepsilon = \left\{ \sigma_i^k \mid B(x_i, \varepsilon/2) \ne \varnothing, \forall x_i \in \sigma_i^k \right\} \quad (2)$$

As evident from the above equation, if the distance between two points in the point cloud is less than $\varepsilon$, they can form a subset of 1-simplex in the Vietoris-Rips complex $V_\varepsilon$. Similarly, three 1-simplex can form a 2-simplex, and so forth. Thus, the Vietoris-Rips complex $V_\varepsilon$ is comprised of all subsets of simplex that satisfy this condition, forming a simplicial complex structure. Clearly, as the distance parameter $\varepsilon$ increases, the original simplicial complex remains unchanged but incorporates new complex corresponding to larger distance parameters. Therefore, in the point cloud $X$, the Vietoris-Rips complex $V_{\varepsilon_i}(X)$ for different distance parameters $\varepsilon_i$ exhibit the following inclusion relationship:

$$V_{\varepsilon_1}(X) \subseteq V_{\varepsilon_2}(X) \subseteq \ldots \subseteq V_{\varepsilon_n}(X) \quad (3)$$

In a given simplicial complex $V_{\varepsilon_i}(X)$ corresponding to a distance parameter $\varepsilon_i$, various topological features exist. In Persistent Homology, global topological features are defined by the Betti numbers $\beta_k$, which also represent the k-dimensional holes $H_k$ in the topological space. Specifically, $\beta_0$ represents the number of connected components, $\beta_1$ represents the number of holes in two-dimensional surfaces, $\beta_2$ represents the number of voids in three-dimensional regions, and so forth. Each simplicial complex corresponds to homological groups $H_k\left(V_{\varepsilon_i}(X)\right)$ with the following relationships:

$$H_k\left(V_{\varepsilon_1}(X)\right) \subseteq H_k\left(V_{\varepsilon_2}(X)\right) \subseteq \ldots \subseteq H_k\left(V_{\varepsilon_n}(X)\right) \quad (4)$$

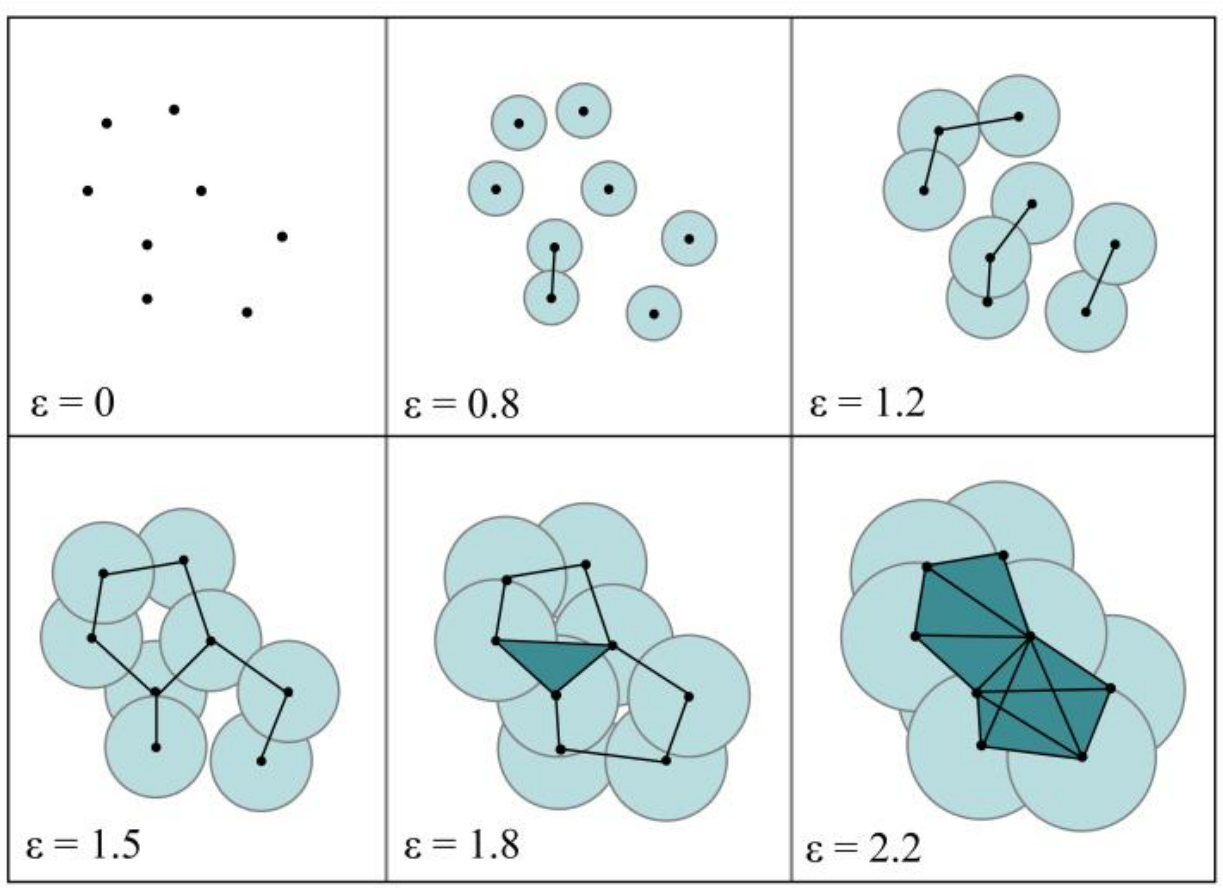

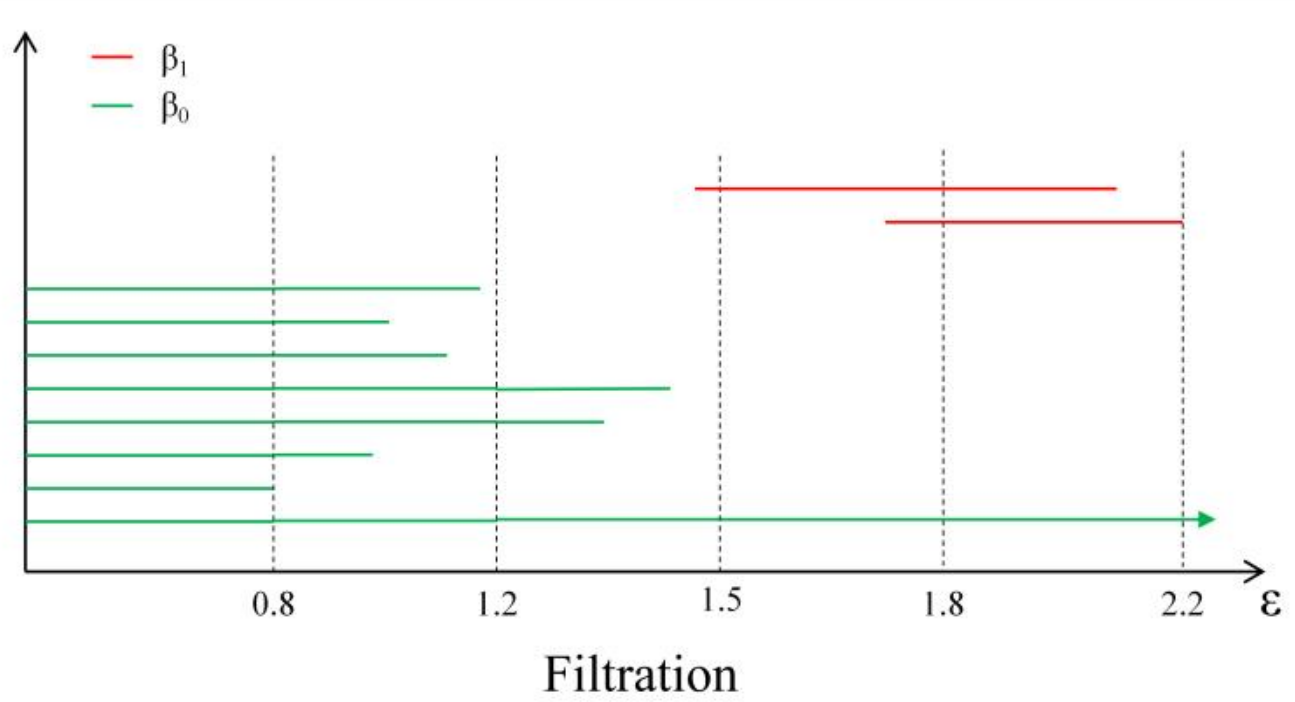

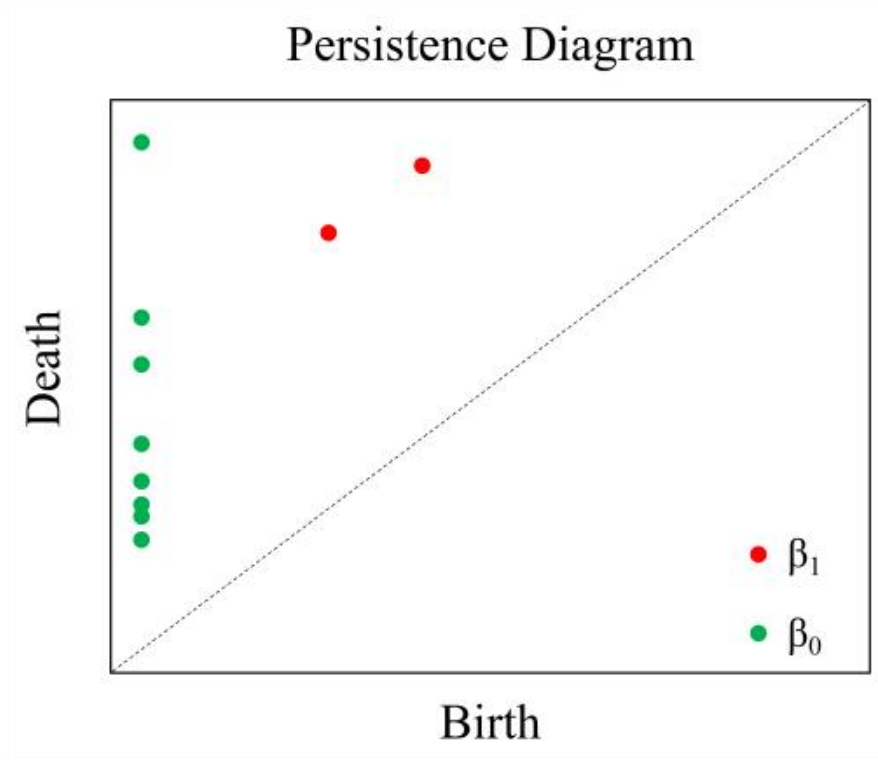

**Figure 3.** From top to bottom: formation of *Filtration* with distance $\varepsilon$ changes in Persistent Homology; Persistence Barcodes formed by recording births and deaths of topological feature $\beta_k$ during Filtration; Persistence Diagram, a derived form of Persistence Barcodes.

During the process of continuously changing $\varepsilon$, the simplicial complex with inclusion relations form a *Filtration*. In the construction of *Filtration*, the generation and extinction of $\beta_k$ occur. The main idea of Persistent Homology is that topological features $\beta_k$, which persist in a large scale, are considered representative topological features through the computation of the dimension of the homological group $H_k\left(V_{\varepsilon_i}(X)\right)$. In TDA, the information of topological features $\beta_k$ is typically recorded in Persistence Barcodes and Persistence Diagrams, as shown in Figure 3. The image accurately reveals the birth and death times of each $\beta_k$ by tracking the variation of the distance parameter $\varepsilon$. As the dimension of $\beta_k$ increases, its computational resources also exhibit exponential growth. Taking into account these factors, we choose $\beta_1$, which best represents the two-dimensional topological features, for computation.

### *3.2. Improvement of TDA Methodology in ADHD EEG*

While the application of Topological Data Analysis (TDA) helps to reveal nonlinear topological features within signals, in the case of EEG signals, the point cloud representations may not always exhibit significant topological structures. This phenomenon has been observed multiple times in the nonlinear analysis of EEG [44-45], where the point cloud construction typically manifests as a scattered distribution of high-density points throughout the phase space, thus obscuring the underlying topological structures as neighboring points continuously disrupt them, leading to the loss of topological features. Fortunately, in TDA, even when the data lacks topological structures, points closest to the diagonal still emerge, albeit often interpreted as noise, posing challenges to our analysis. Although the point cloud construction of ADHD's EEG contains relatively small topological features, TDA fails to extract them. Instead, in the Persistence Diagram, excessively short birth-death times overshadow these features and hide them under topological noise.

This deficiency becomes even more severe after multi-channel mapping, as depicted in Figure 4. Even if the point cloud formed in certain channels contains sufficiently large topological structures, other channels may lack corresponding features or their feature structures may not fall within a certain range. After remapping, the final point cloud fails to enhance this topological structure and may even cause features to disappear. Even if the Persistence Diagram in individual channels displays topological features, the overall mapped Persistence Diagram will only exhibit topological forms filled with noise, resulting in the loss of topological features.

Avoiding this feature loss phenomenon is the focus of our research. We have found that after phase space reconstruction, the hidden topological structures exhibit density differences compared to typical phase points. This suggests that we can analogize the topological recovery methods for dealing with high noise impact in TDA with topological feature search. This means that we can use derived methods of the distance-to-measure (DTM) function to extract the original features, specifically employing the k-power-distance-to-measure (k-PDTM) [46] to achieve the separation of topological structures.

In the DTM function, the mapping of any point $x$ in a point cloud $\mathcal{Z}$ containing $n$ points can be represented in the following form:

$$d_{\mathcal{Z},q}^2 : x \rightarrow \inf_{c \in \mathrm{R}^d} \left\| x - m(x, \mathcal{Z}, q) \right\|^2 + v(x, \mathcal{Z}, q) \quad (5)$$

Where $m(x, \mathcal{Z}, q) = \frac{1}{q}\sum_{i=1}^{q} X^{(i)}$ , $X^{(i)}$ represents the $i$-th nearest neighbor point of $x$ (for $i = 1, 2, \ldots, q$ ), and $m(x, \mathcal{Z}, q)$ denotes the centroid of $q$ nearest neighbor points of $x$ . $v(x, \mathcal{Z}, q)$ represents their variance, that is $v(x, \mathcal{Z}, q) = \frac{1}{q}\sum_{i=1}^{q} \left\| m(x, \mathcal{Z}, q) - X^{(i)} \right\|^2$ .

However, computing the joint homology of $n$-balls for DTM sublevel sets would consume significant computational resources in this experiment. Therefore, we employ approximate sublevel sets of k-PDTM as an approximation for $n$ points. In k-PDTM, only the joint homology of $k$-balls is needed to approximate the original $n$-balls, computed as follows:

$$d_{\mathcal{Z},q,k}^2 : x \rightarrow \min_{i \in \{1,2,\ldots,k\}} \left\| x - m(c_i^*, \mathcal{Z}, q) \right\|^2 + v(c_i^*, \mathcal{Z}, q) \quad (6)$$

where $c_i^*$ represents the geometric centres of the $k$-approximation balls, which are determined by the following equation:

$$(c_1, c_2, \cdots, c_k) \rightarrow \sum_{X \in \mathcal{Z}} \min_{i \in \{1,2,\ldots,k\}} \| X - m(c_i, \mathcal{Z}, q) \|^2 + v(c_i, \mathcal{Z}, q) \quad (7)$$

As shown in Figure 5, we select 350 optimal centres in the phase space mapping of ADHD, and these $k$-balls centres are selected by Equation 7.

After the computation with k-PDTM, $\forall x \in \mathcal{Z} \subseteq \mathrm{R}^d$ is assigned a distance measures. Figure 5 demonstrates that the distance measures obtained using $k$-balls sublevel sets selected by k-PDTM and those obtained using sublevel sets with $n$-balls are similar, proving that k-PDTM maintains its performance with fewer computational resources. We base topological feature extraction on these distance measures.

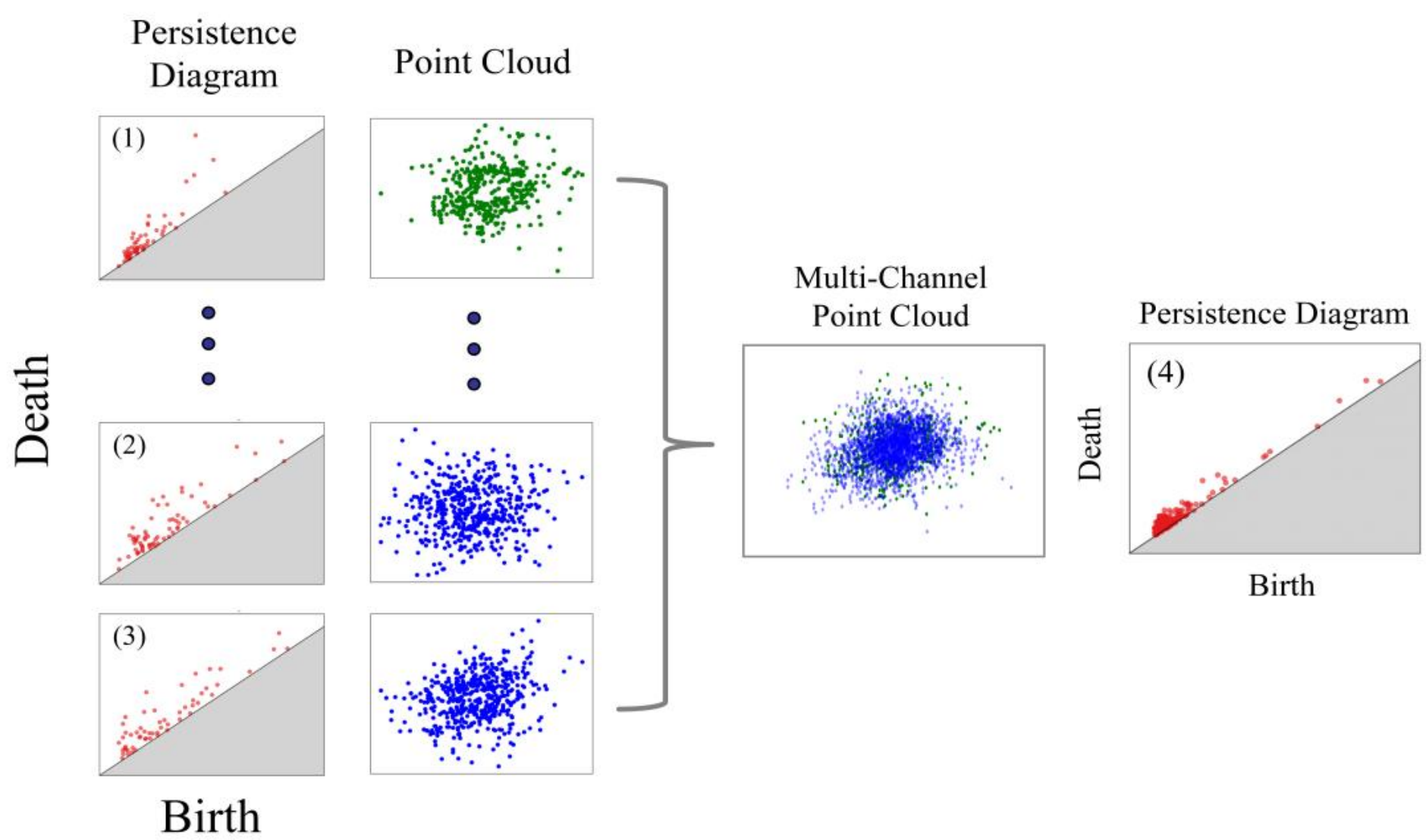

**Figure 4.** The phenomenon of topological feature loss in 2D point clouds. (1). A point cloud containing topological features whose features are then revealed in a Persistence Diagram. (2)-(3). Point cloud with topological features covered, Persistence Diagram contains only topological noise (4). Point cloud after remapping with Persistence Diagram, topological features in (1) are lost.

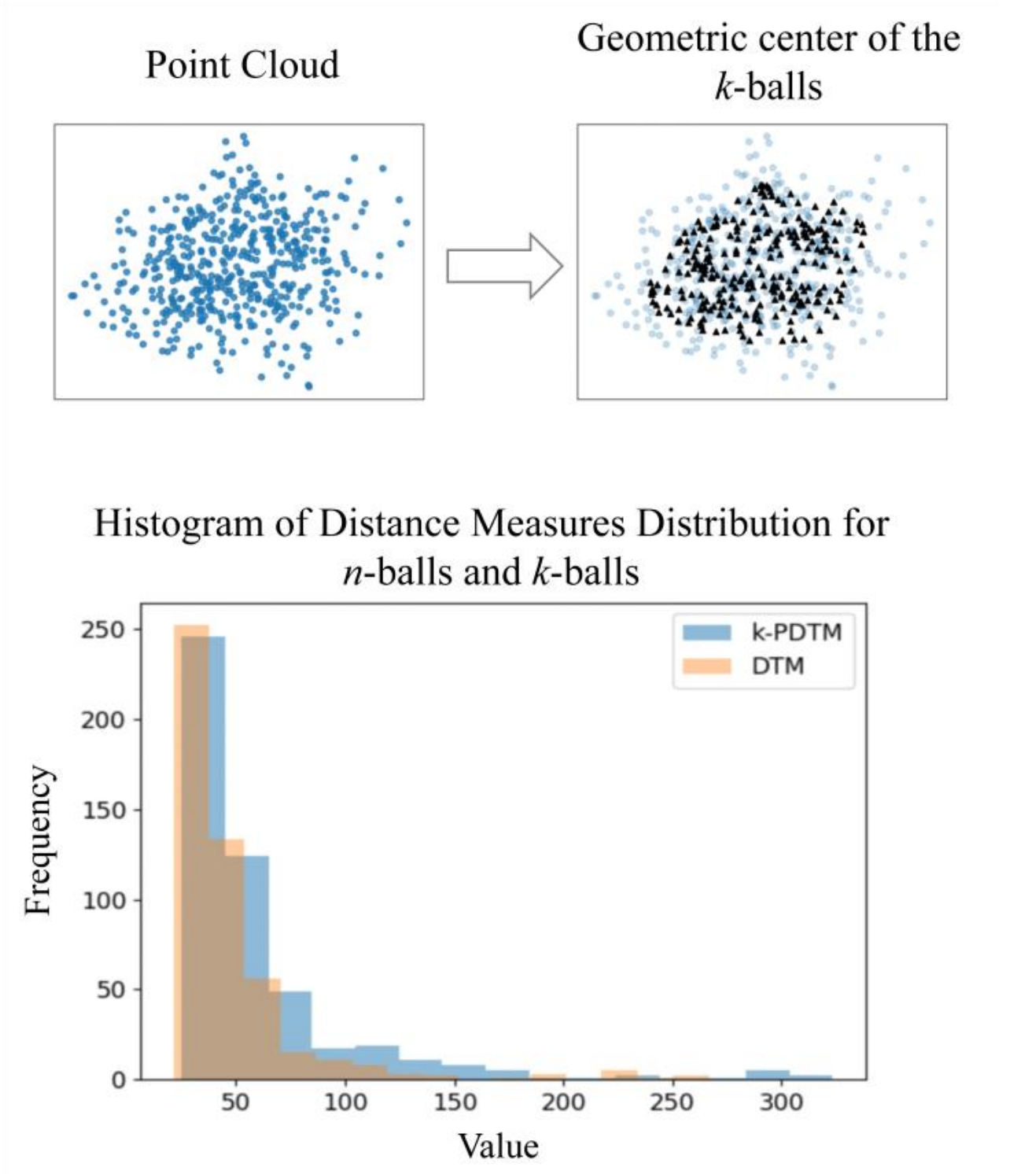

**Figure 5.** From top to bottom: the geometric centre of the $k$-balls is confirmed using the k-PDTM; the histogram of the distance measures distribution obtained from the k-PDTM and the DTM in the same point cloud shows that the trend of the k-PDTM values is similar to that of the DTM.

In ADHD, the repetitive mapping of high-density points in the phase space disrupts the original topological structure. Therefore, we traverse the distance measures of all points, where smaller values, representing proximity to the geometric center of the $k$-approximate ball, indicate closer proximity to a compact set and thus greater disruption to the topological structure. These points should be removed from the sample points, contrary to their application in robust topological inference. After multiple experiments, we determined that retaining 140 phase points for each single-channel point cloud is the optimal parameter in multivariate analysis remapping. Our method not only extracts topological features from topological noise in persistence plots but also does not affect channels with already significant topological structures in multivariate analysis, preserving their mappings in the full phase space in the persistence plots.

### *3.3. Filtering of Persistence Diagram*

For long-term EEG recordings, simple slicing of EEG data may result in unbalanced data segments [47]. Therefore, we decided to utilize all EEG slices for ADHD assessment. In the previous section, we extracted the topological features of individual slices in persistence diagrams. In this section, we use the collection of persistence diagrams to represent the entire topological features of a single ADHD patient.

However, considering the influence of extreme points and noise, in the superposition of persistence diagrams, points with fewer occurrences within a unit birth-death interval may arise from unbalanced segments. In ADHD EEG, the selected topological features exhibit convergence, with their mappings in the persistence diagram clustering within specified intervals. Therefore, we can extract the birth-death points from the persistence diagram onto the $R^2$ and then utilize Multivariate Kernel Density Estimation (MKDE) to assess the relative density of points. Subsequently, we filter out points with lower values, separate outliers, and remap the remaining points back to the persistence diagram as the final topological features of ADHD. The MKDE function is defined as follows:

$$f(x) = \frac{1}{N}\sum_{i=1}^{N} K(x - x_i) \tag{8}$$

In the above equation, $K$ represents the multidimensional kernel function, and $x_i = \left[x_i^{(1)}, x_i^{(2)}, \ldots, x_i^{(n)}\right]^T$. In selecting $K$, we opt for the most stable Gaussian kernel as the kernel function for our experiments. Its multidimensional form is expressed as:

$$K = (2\pi)^{-\frac{m}{2}} \det(H)^{-\frac{1}{2}} e^{-\frac{1}{2}x^T H^{-1} x} \tag{9}$$

In the equation above, $H$ represents the bandwidth matrix, defined as $H = 10M$. Following the application of MKDE, since the number of points in each persistence diagram varies, we opt for a proportional threshold rather than a fixed one. Specifically, we arrange all points in ascending order based on their values, then select the top 99% of points as the threshold, discarding points with values greater than this threshold. Finally, we remap the points from the $\mathbb{R}^2$ back to the persistence diagram.

### *3.4. Topological Features Extraction*

In TDA, the obtained persistence diagrams cannot be effectively embedded in various machine learning classifiers. Therefore, it is necessary to transform the persistence diagrams and further extract the contained topological features [48]. Among various methods, Entropy Summary Function (ES) [49], Persistence Landscape (PL) [48], and Persistence Image (PI) [50] are commonly used for feature extraction. They are capable of mapping barcodes and persistence diagrams into elements of vectors, enabling statistical analysis and the establishment of machine learning models. However, ES and PL methods lack flexibility in the process of topological feature extraction. The fixed mapping approach prevents us from fully experimenting with the differences in the generation of feature vectors for each persistence point. Several studies have shown that medium and small persistence points can affect classification results [51-52]. The PI method addresses this issue by providing selectable weighting functions, allowing for the adjustment of the proportion of various types of persistence points in the mapping. This improvement undoubtedly offers advantages over the other two methods of analysis. Therefore, we choose the PI method as the approach for extracting topological features from persistence images.

In the PI method, the first step is to linearly map the points on the persistence diagram to a two-dimensional plane, denoted as $T:\mathrm{R}^2 \to \mathrm{R}^2$ . We choose the default linear function $T(x,y)=(x,y-x)$ for this mapping. Let $B$ represent the birth-death coordinates in the persistence diagram. Then, $\rho_B:\mathrm{R}^2 \to \mathrm{R}$ maps it to the persistence surface in the Persistence Image as follows:

$$\rho_B(z)=\sum_{u\in T(B)} f(u)\phi_u(z) \tag{10}$$

Where $z=(x,y)$ , and $\phi$ represents the standardized symmetric Gaussian probability distribution, $\phi_u(x,y)=(2\pi\sigma^2)^{-1}e^{-\left[(x-u_x)^2+(y-u_y)^2\right]/2\sigma^2}$ , Where $u$ and $c$ represent the mean and variance respectively, while $f(u)$ is defined as a non-negative weighted function along the horizontal axis, which is continuous and piecewise differentiable. It adjusts the impact of each persistence point in the Persistence Image. The final Persistence Image is a collection of pixels $I(\rho_B)_P=\iint_P \rho_B \; dydx$ .

In PI, the weighting function typically depends only on the vertical persistence coordinate $y$ , i.e., $f(x,y)=w_b(y)$ . Common choices for the weighting function include non-decreasing sigmoidal piecewise functions or $y=x^2$ , which helps to maintain a balance between low-persistence points and high-persistence points to some extent. we aim to comprehensively analyze the influence of both low-persistence and high-persistence points on the topological features of ADHD EEG in PI. Therefore, our weighting function is constructed as follows:

$$w_b(y)=\begin{cases} a & 0<y\le t_1 \\ \dfrac{c-a}{t_2-t_1}y+\dfrac{t_2a-t_1c}{t_2-t_1} & t_1<y\le t_2 \\ (y-t_2)^2+c & y>t_2 \end{cases} \tag{11}$$

The function satisfies the condition of being zero along the horizontal axis, continuous, and piecewise differentiable. By altering the parameters of this function, we can conduct comprehensive experiments on the persistence points. Firstly, we need to determine the values of certain parameters in Equation 11 based on the characteristics of topological features in ADHD. This is to avoid introducing too many variables in the subsequent processes.

As shown in Figure 6, the mapping of most persistence points in the experiment has their $y$ values in $\mathrm{R}^2$ within 100. In our method, large-scale mapping can lead to an increase in the number of low-persistence points. According to Equation 10, their Gaussian distribution will be larger than that of high-persistence points. Therefore, we set the weighting function for low-persistence points to a constant value, aiming to balance the influence of their increased quantity. Meanwhile, we set $t_1=100$ to ensure that this weighting does not affect high-persistence points. For high-persistence points, they are generally distributed within 100-200. Thus, we set $t_2=200$ and provide them with a linearly increasing weighting to compensate for the disadvantage in their quantity. Furthermore, there are fewer extremely high-persistence points distributed beyond 200. For these points, we assign an exponentially increasing weighting to enhance the pixel intensity in their Persistence Image representation, finally, the pixels has been obtained .

Although it is a standard practice to use PI images to reorganize the entire PI image into rows or columns into a set of one-dimensional features, after using the weighting function, a large number of irrelevant variables in the image will affect the final classification result. Therefore, we directly use the pixel with the largest intensity in the image as the feature for classification. In our experiment, this feature retains most of the information of the PI image.

After the above analysis, we have determined the parameters $t_1$ and $t_2$ . Now, among the variables in the weighting function parameters, only $a$ and $c$ remain to be defined. Here, $a$ alters the influence of low-persistence points on the Persistence Image, while $c$ modifies the impact of extremely high-persistence points on the Persistence Image. The parameters $a$ and $c$ together define the effect of high-persistence transition points on the Persistence Image.

As depicted in Figure 7, within the PI method, we can extract different topological features from the same persistence diagram by adjusting $a$ and $c$.

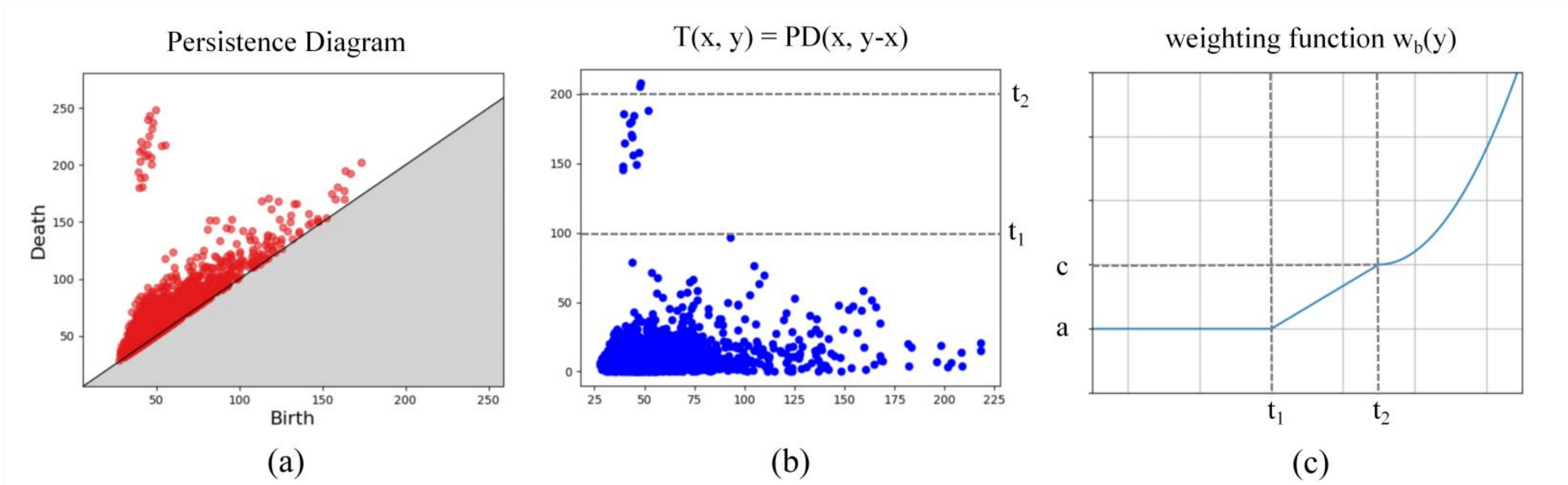

**Figure 6.** (a). Persistence Diagram obtained after the preliminary steps. (b). Mapping the persistence points in the PD to $\mathrm{R}^2$ using the linear function $T(x, y)$. (c). Construction of the weighting function $w_b(y)$.

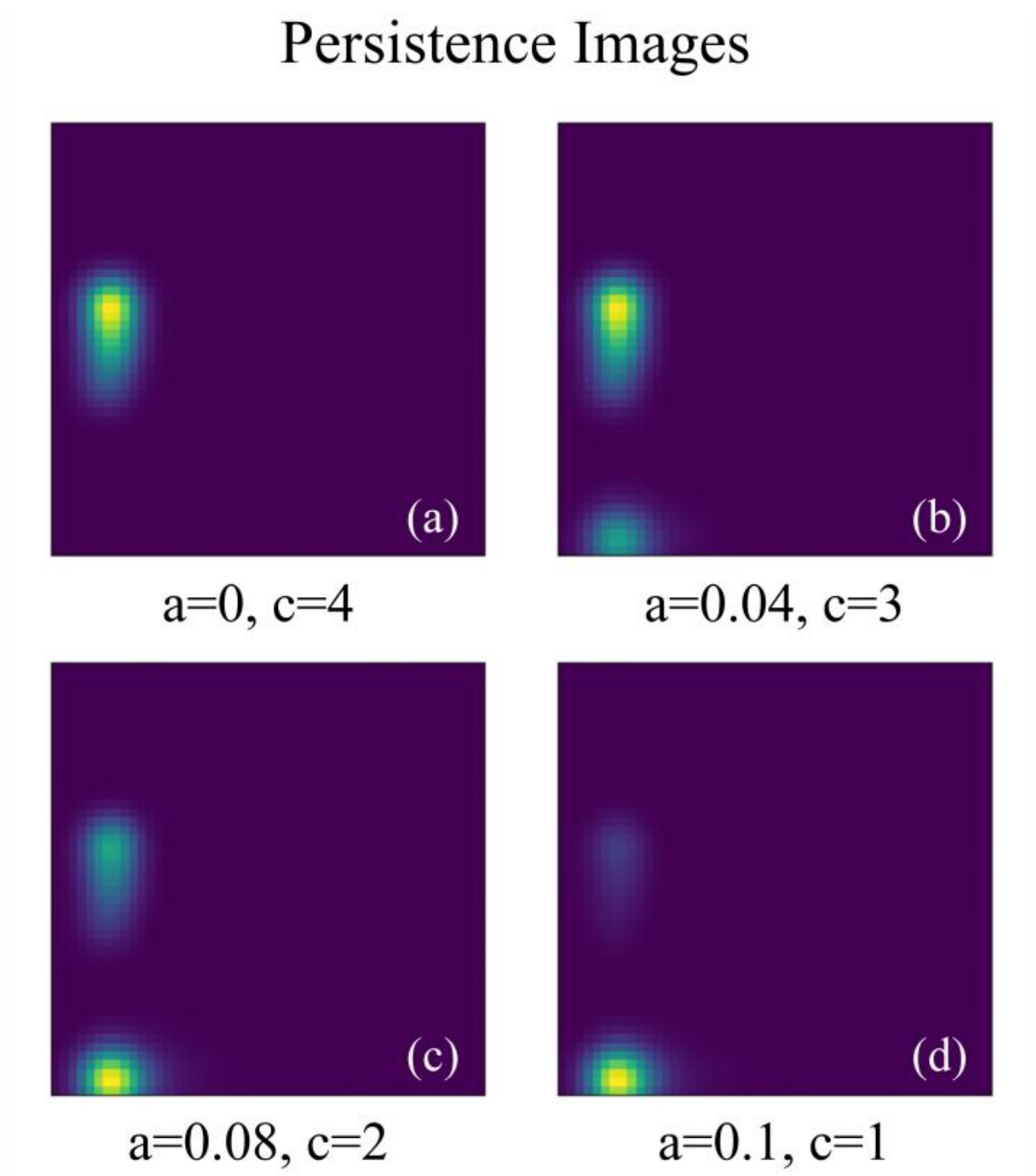

**Figure 7.** (a). Persistence images using weighting functions with different parameters. Parameter $a$ has a significant impact on low persistence points, while parameter $c$ has a significant impact on high persistence points.

## 4. Experiment and Discussion

In this section, we will embed the features extracted from the persistence images obtained in the previous section into machine learning classifiers and classify individuals as either ADHD patients or healthy controls. Firstly, we will present the machine learning classifiers and evaluation metrics used. Subsequently, we will conduct multiple cross-validation experiments and showcase the results, followed by a discussion on the reasons behind these results. Finally, we will present and compare the outcomes obtained from our method with those from other approaches.

### *4.1. The relevant evaluation metrics*

To assess the effectiveness of the proposed algorithm, we adopted quantitative evaluation methods, specifically utilizing three evaluation metrics: Accuracy (ACC), Sensitivity (SE), and Specificity (SP). These metrics comprehensively evaluate the performance of the algorithm in extracting topological features in the classifier [53].

Accuracy is the most common intuitive statistical measure, applicable to all samples. It specifically refers to the proportion of correctly classified samples out of the total number of samples, expressed as follows:

$$ACC = \frac{TP + TN}{TP + FN + FP + TN} \tag{12}$$

Where $TP$ stands for True Positive, indicating cases where the predicted value is positive and the true value is also positive; $FN$ stands for False Negative, indicating cases where the predicted value is negative and the true value is positive; $FP$ stands for False Positive, indicating cases where the predicted value is positive and the true value is negative; and $TN$ stands for True Negative, indicating cases where the predicted value is negative and the true value is also negative.

Although the metric of accuracy is simple and effective, it does not reflect the details of the classification, so we added sensitivity and specificity, and their formulas are as follows:

$$SE = \frac{TP}{TP + FN} \tag{13}$$

$$SP = \frac{TN}{FP + TN} \tag{14}$$

As indicated by the above formulas, sensitivity reflects the proportion of true positive instances correctly classified by the classifier among all positive instances, while specificity reflects the proportion of true negative instances correctly classified by the classifier among all negative instances. By considering all three metrics together, we can more comprehensively assess the effectiveness of the topological feature descriptor in classification.

In selecting a machine learning classifier, we plan to use the K Nearest Neighbour algorithm (KNN) for feature classification. The reason for choosing KNN is that KNN has shown superior performance in classifying individual features in small samples compared to other classifiers [54].

### *4.2. The Experimental Results and Comparison*

*1). Selection of optimal PI parameters:* In Section 3.4, we employed the Persistence Image method to further extract topological features from the obtained persistence diagrams. In the design of the final weighting function, we retained two parameters $a$ and $c$ that affect the points with lower persistence, close to the diagonal (topological noise) and higher persistence, away from the diagonal (topological features) respectively. In this section, we experiment with combinations of these two parameters to evaluate their impact on the overall classification results, thereby determining the optimal weighting function of the PI method for ADHD.

As shown in Figure 8, we conducted experiments using four sets of data, among which the parameter combination of $c = 3$ and $a = 0$ achieved the highest accuracy, sensitivity, and specificity. The performance obtained with this parameter combination far exceeded that of other parameters. Therefore, we decided to adopt the parameter combination of $c = 3$ and $a = 0$ to refine the weighting function set in our PI method, ensuring its optimal performance in ADHD classification tasks.

Combining the experimental observations above, we can conclude that in ADHD analysis, the presence of lower persistence points leads to a decrease in classification accuracy. This phenomenon arises from the increased persistence of topological noise due to repeated mappings in multi-channel analysis, ultimately resulting in a significantly higher number of lower persistence points compared to higher persistence points. Therefore, in the PI method, the Gaussian probability distribution upon which pixel value calculations depend offsets the advantages brought by the weighting function, causing pixels with higher intensities to fall on lower persistence points. However, there is no significant distinction in the topological features of ADHD EEG and healthy control EEG in terms of lower persistence, which is not conducive to further classification. This experiment indicates that the presence of lower persistence points is not necessary in multi-channel analysis of ADHD. On the other hand, higher values of $c$ may cause outliers that are not completely separated by MKDE to affect the pixel Intensity, leading to a decrease in classification accuracy.

*2). Comparison with other topology descriptors:* In Section 3.4, we explained why we chose the Persistence Image method over other topological feature extraction methods to obtain the final topological descriptors. In this subsection, we aim to demonstrate this result more intuitively through evaluation metrics. Firstly, we obtained persistence diagrams using our method. Then, we computed feature vectors using the Entropy Summary Function, Persistence Landscape, Betti number, and Persistence Image separately. After obtaining these feature vectors, we conducted classification experiments using KNN. Table 1 represents the classification results.

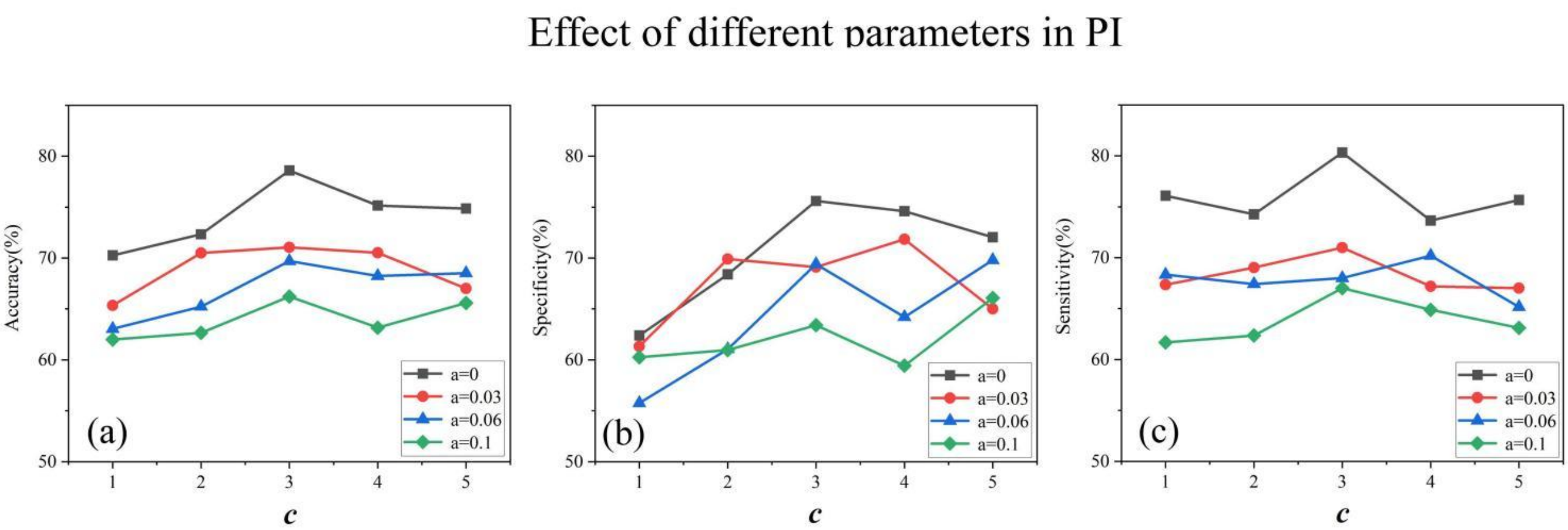

**Figure 8**. The experiment on the impact of various parameters on classification accuracy shows that as the parameter $a$ decreases, the accuracy, specificity, and sensitivity gradually increase. (a). The effect of parameter on accuracy. The highest accuracy is 78.27% at $a = 0$ and $c = 3$, while the lowest accuracy is 60.97% at $a = 0.1$ and $c = 1$ . (b). The effect of parameter on specificity. The highest specificity is 75.63% at $a = 0$ and $c = 3$ , while the lowest specificity is 55.75% at $a = 0.06$ and $c = 1$ . (c) The effect of parameter on sensitivity. The highest sensitivity is 80.62% at $a = 0$ and $c = 3$ , while the lowest sensitivity is 61.65% at $a = 0.1$ and $c = 1$ .

The results indicate that the PI method, with the configured weighting function, outperforms other methods in terms of accuracy, sensitivity, and specificity, achieving 78.27%, 80.62%, and 75.63%, respectively. Among other topological feature descriptors, Persistence Landscape shows better performance, with accuracy, sensitivity, and specificity reaching 74.38%, 71.67%, and 77.05%, respectively. However, using Betti numbers as topological feature descriptors shows no discrimination at all, which indirectly indicates that the topological features derived from ADHD EEG are based on persistence time rather than generation quantity, reaffirming the analysis in 1) that low persistence points have a significant impact on the accuracy of classification results. Overall, the PI method as a topological descriptor has the advantage of flexibility. It can adapt to different PD images and provide the optimal solution based on its own parameter settings. Its comprehensive performance is generally higher than other topological descriptors by more than 3-5%.

*3). Comparison with original TDA:* Through experiments, we can observe the improvements of the method proposed in this paper compared to the original TDA method. Before section 3.4, we only obtained persistence diagrams without further extracting their topological features. This means that at this time, our method and the TDA method both yield birth-death points that are not influenced by subsequent processing methods. The persistence diagram is a side manifestation of the nonlinear topological features of the signal point cloud. To avoid the influence of the weighting function we set on the PI method, we only used persistence landscapes and Entropy Summary Function for comparison. Additionally, because the original TDA method cannot perform multi-channel calculations, we thoroughly examined all EEG channels used. As shown in Figure 9, we applied the original TDA method to all channels and used persistence landscapes and Entropy Summary Function to extract topological features.

Table 1. Experiments using individual topology descriptors

| Topological Descriptors | Accuracy (%) | Specificity (%) | Sensitivity (%) |
|---|---|---|---|
| Entropy | 71.07 | 72.13 | 70.00 |
| Landscape | 74.38 | 77.05 | 71.67 |
| Betti | 53.72 | 56.67 | 50.81 |
| PI | **78.27** | **75.63** | **80.62** |

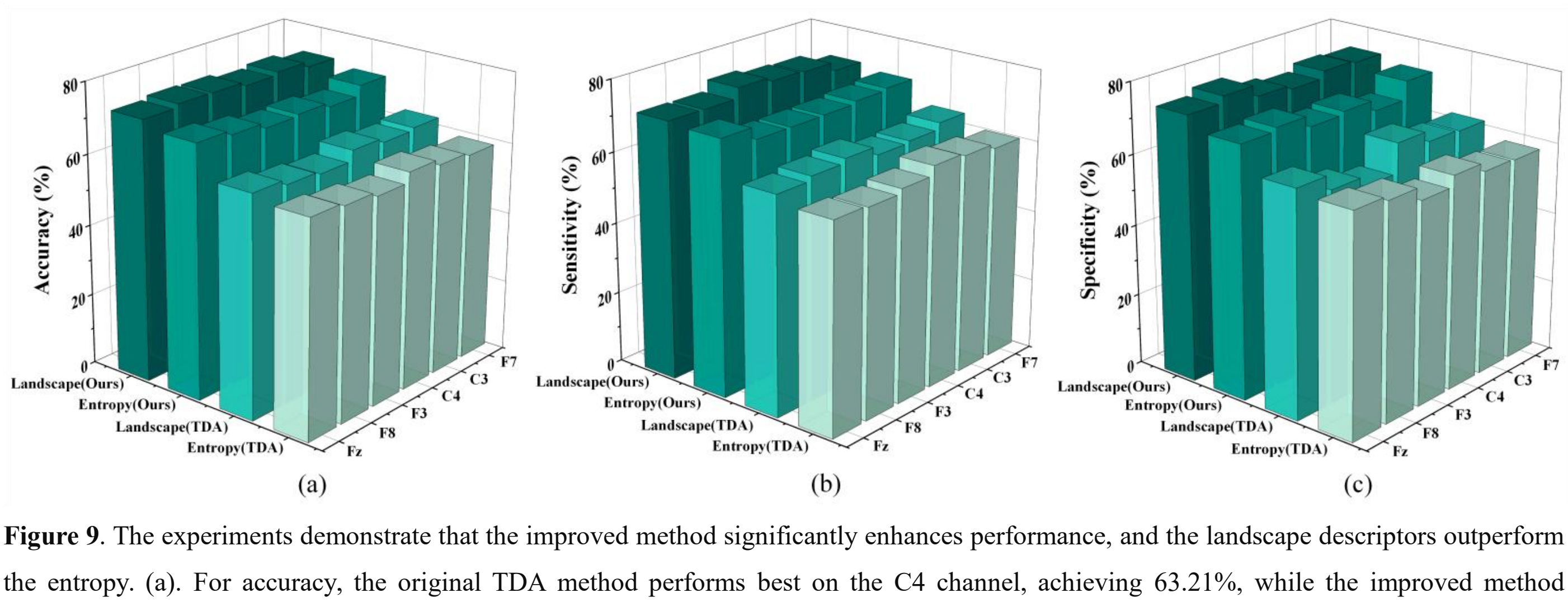

**Figure 9**. The experiments demonstrate that the improved method significantly enhances performance, and the landscape descriptors outperform the entropy. (a). For accuracy, the original TDA method performs best on the C4 channel, achieving 63.21%, while the improved method

performs best on the F8 channel, achieving 73.29%. (b). For sensitivity, both the original and improved methods perform best on the F3 channel, with sensitivity of 63.52% and 75.14%, respectively. (c). For specificity, the original TDA method performs best on the C4 channel, achieving 65.20%, while the improved method performs best on the F8 channel, achieving 76.65%.

The performance of both sets of data can be seen, it can be observed that the features extracted from the signal point cloud by the original TDA method lack separability. The reason for this phenomenon is well understood: in the original TDA method, the phenomenon of losing topological features mentioned in section 3.2 causes all persistence points to converge towards the diagonal line. Whether it is ADHD EEG or healthy population EEG, their birth-death values are relatively close, making it difficult to achieve effective differentiation using general statistical measures. In contrast, our method delays the death time of the persistence point set. The topological features of ADHD EEG disappear later compared to those of healthy population EEG, resulting in a higher average death value of persistence points in ADHD. Additionally, from Formula 4, it can be seen that the newly generated homology group contains the old homology group. Therefore, during the *Filtration* process, the number of smaller persistence points is much larger than that of larger persistence points. This indicates that the overall performance of the persistence points is relatively stable, whether it is ADHD EEG or healthy population EEG, leading to the poor performance of other descriptors mentioned in 2).

4). *Statistical tests:*

- Prior to feature extraction, it is necessary to validate the trends observed in the data and ensure the robustness of the analysis, by identifying and testing these properties we aim to provide more reliable and interpretable results. Firstly, we use time series decomposition to decompose the original signal into trend, period and residuals, observing the trend of the signal to initially determine whether the signal is non-stationary or not, followed by fitting the residuals of the signal using an ARCH (Autoregressive Conditional Heteroskedasticity) model, which assumes that the variance at the current moment is related to the square of the residuals of the past few moments only, and the use of ARCH allows for a better check of the time series' heteroscedasticity [55].

We used the typical signal channel of ADHD (frontal lobe, Fz) and downsampled the signal to make the analysis easier, typical results using time series decomposition and ARCH are shown in Figure 10.

We analysed all ADHD patients, and the trend curve of the time series decomposition can initially show the non-stationary trend of their EEG signals, and furthermore, in the ARCH model, Standardized Residuals represent the volatility characteristics of the time series captured by the ARCH model, while the Conditional Volatility obtained from the fitting result fluctuates greatly, indicating that the signal has heteroscedasticity.

Lastly, we aimed to verify whether this type of EEG signal represents a fractal-like model, specifically a fractional Brownian motion. If the EEG exhibits such characteristics, it would possess attractors and fractal structures in phase space, confirming the suitability of our feature extraction method. To validate this, we used the Hurst exponent [56]. If the Hurst exponent $H$ falls within the range of 0 to 0.5, the time series data can be described as a biased Brownian motion (fractional Brownian motion). The experimental data is shown in Figure 11.

The results show that the experimental EEG can be described as a fractal Brownian motion, i.e. the method based on phase space mapping with TDA feature extraction used in the paper is applicable.

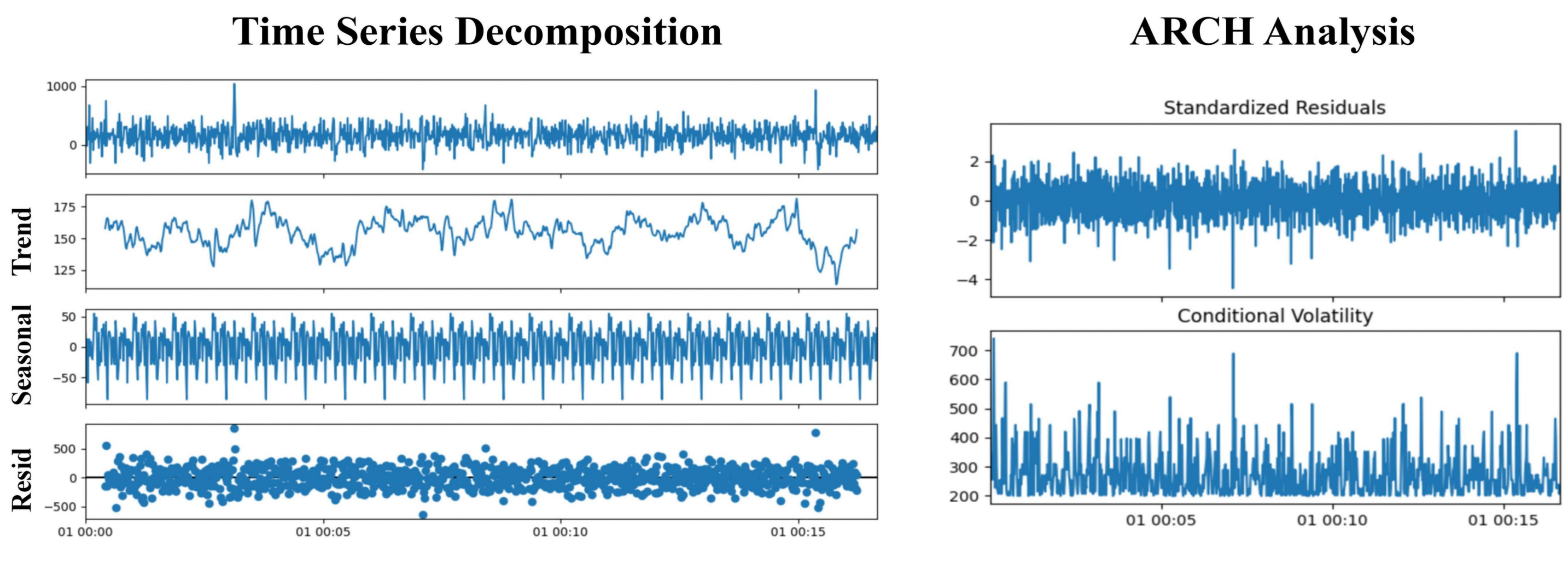

**Figure 10**. Using time series decomposition and ARCH to verify the non-stationarity and heteroscedasticity of EEG signals

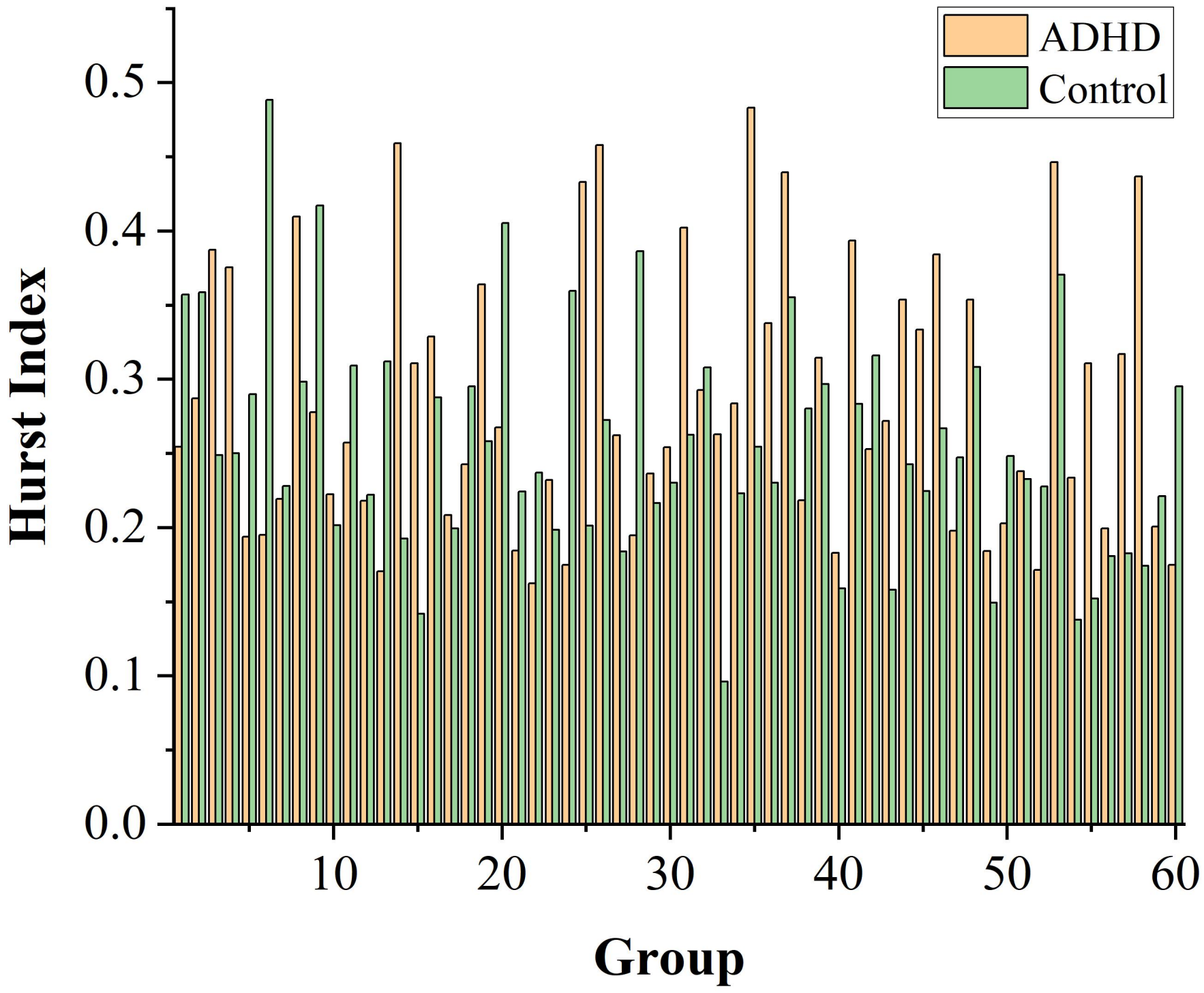

**Figure 11**. Verification of EEG for fractal properties using the Hurst index

- In order to assess whether the extracted features have the ability to distinguish between the datasets independently of the machine learning algorithms, we conducted statistical tests using the most suitable feature parameters determined in the previous section. We selected three methods for this evaluation. First, we applied an independent samples t-test, which assumes that our feature data follows a normal distribution. Secondly, we used the Mann-Whitney U test, a non-parametric test that assesses whether two samples come from the same distribution and can detect significant differences between them. It does not rely on the normality assumption and can also mitigate the influence of outliers. Finally, to avoid the impact of numerical differences and sample size on the results, we calculated Cohen's d to quantify the effect size between the two groups [57]. The formula for Cohen's d is as follows:

$$d = \frac{\overline{X_1} - \overline{X_2}}{s_p} \tag{15}$$

Among them, $s_p$ is the combined value of the standard deviation of the two groups, and the formula is:

$$s_p = \sqrt{\frac{(n_1 - 1)s_1^{\ 2} + (n_2 - 1)s_2^{\ 2}}{n_1 + n_1 - 2}} \tag{16}$$

In the context of Cohen's d values, an effect size around 0.2 represents a small effect, around 0.5 represents a medium effect, and around 0.8 represents a large effect. A larger effect size indicates a greater difference between the two groups, supporting the effectiveness of the feature in classification.

Our test results are shown in Table 2. The results demonstrate that the extracted features exhibit significant differences across different datasets, and the quantified effect sizes are substantial. These findings confirm that the extracted features possess strong discriminative power and can effectively distinguish between different conditions in the datasets, independently of machine learning methods.

5). *Comparison of related work:* Currently, various feature extraction methods have been proposed for ADHD classification in research studies. The focus of this paper is to propose a new nonlinear descriptor to describe the topological features of ADHD EEG for efficient classification.

- **Entropy Feature:** Among other nonlinear descriptors, Nassehi [58] used approximate entropy (ApEn), Petrosian fractal dimension (Petrosian), and Lyapunov exponent (LLE) to assess the performance of nonlinear features; Rezaeezadeh [59] employed nonlinear entropy features such as Shannon entropy (ShanEn), sample entropy (SampEn), dispersion entropy (DispEn) and Multiscale sample entropy (MSE) for extensive evaluation.

Table 2. p-values and difference quantification values obtained using three statistical test methods

| Statistical indicators | t-test | Mann-Whitney U test | Cohen's d |
|---|---|---|---|
| Effect index | none | none | 0.706 |
| $p$ | 0.009 ($p$<0.05) | 0.003 ($p$<0.05) | none |

In our comparison, we selected various entropy measures because they play a critical role in nonlinear analysis by capturing the complexity and unpredictability of EEG signals. Entropy provides a quantitative measure of disorder or randomness within a system. In the evolution of entropy concepts, S.M. Pincus first introduced Approximate Entropy (ApEn) [60], which quantifies the complexity of a time series by measuring the likelihood of new patterns emerging as the dimension increases. However, ApEn includes comparisons with self-vectors, so Joshua S. proposed Sample Entropy (SampEn) [61], which reduces the dependence on data length by eliminating the effect of self-matching and offers better consistency compared to ApEn. Furthermore, both ApEn and SampEn calculate complexity on a single time scale, potentially overlooking long-term structures or low-frequency components of the signal, which can be important in complex physiological signals. Costa M. addressed this issue by introducing Multiscale Entropy (MSE) [62], which constructs time series at different scales by downsampling the original sequence, then calculates SampEn or other entropy measures at each scale to estimate complexity across multiple scales. This approach captures the dynamic characteristics of signals over various time scales and reflects the complexity of signals in both coarse-grained (long time scale) and fine-grained (short time scale) views, which cannot be revealed by single-scale ApEn and SampEn.

On the other hand, since EEG is a long time-series, its complexity across different time segments and scales also needs comprehensive analysis. To address this, Ricardo Zavala-Yoe et al. proposed Bivariate Multiscale Entropy (BMSE) [63]. BMSE constructs a complex three-dimensional surface by calculating the MSE curve for each sub-time vector, which encompasses the complexity information of sequences across different time vectors and scales, allowing for the simultaneous study of large-scale, long-term EEG. Additionally, Ricardo Zavala-Yoe defined the BMSE Index (DBMSE) to quantify the complexity states of different brain regions and explore the dynamic activation of these regions over time during various tasks through complexity paths. BMSE has already been applied to epilepsy diagnosis [64] and brain-computer interfaces [65].

Although various entropy measures address the quantitative representation of signal complexity, they are fundamentally based on the statistical properties of time series, primarily describing the complexity of the signal's temporal dynamics. However, the complexity of EEG signals is not only manifested in their temporal dynamics but also in their geometric structure in high-dimensional space. We have demonstrated through the Hurst index that the EEG signals in this study exhibit characteristics of biased Brownian motion, indicating the presence of attractors and fractal structures in phase space (Figure 11). Using persistent homology, researchers can identify topological structures such as connectivity in EEG signals, reflecting periodicity or cyclicity at specific scales in phase space. While entropy algorithms can quantify complexity, they lack geometric interpretation and cannot provide an intuitive representation of the signal in high-dimensional space. Therefore, in our research, TDA offers a unique advantage in capturing and quantifying the global geometric features and spatial distribution of the signals.

In parameter selection of various entropy features, ShanEn serves as a basic metric, quantifying the randomness and complexity of a system merely through the probability distribution of the time series [18]. In ShanEn calculations, the interval selection follows Sturges' rule [66], where $n = 1 + \log_2 N$ , with $N$ being the number of sampling points. Pincus et al. [67] demonstrated that when the embedding dimension $m = 2$ and the similarity threshold $r = 0.25SD(X)$ ( $SD(X)$ being the standard deviation of the time series), both ApEn and SampEn exhibit reasonable statistical properties. On the other hand, DispEn extends ShanEn by incorporating a discretization mapping step, making it more suitable for complex data and faster to compute [68]. Rostaghi et al. [69] mentioned that for EEG signals, an appropriate parameter selection is embedding dimension $m = 2$ and classes $c = 4$ .

Finally, the calculation of both MSE and CMSE is based on the mean Sample Entropy across different coarse-grained sequences. Thus, the embedding dimension $m = 2$ and similarity tolerance threshold $r = 0.25SD(X)$ are still used, with calculations applied across 20 different time scales.

It is worth mentioning that, considering we have already selected the relevant channels based on the literature [30], we specifically chose the most relevant electrodes for ADHD patients in the experiment. This was done to enhance the geometric characteristics of EEG signals (for both ADHD patients and the control group), making our method more effective. The selected electrodes ensured that the time series from different channels had strong correlations. Therefore, when selecting entropy features, we did not include features that quantify the dynamic interdependence between multiple time series, such as Multivariate Multiscale Entropy (MMSE) and Dynamic Bivariate Entropy (DBE), which are typically used to quantify the association and synchronization between dynamic systems [70].

- **Other Features:** On the other hand, research on time-frequency domain features and related combination features is also common in ADHD classification. For example, Parashar [71] selected comprehensive regional connectivity features by linearly combining the EEG signals from the left hemisphere, right hemisphere, and whole-brain regions, and then input these combined signals into a machine learning classifier. Holker [72] extracted 2424 features from five frequency bands of the raw EEG signals, including spectral, amplitude, functional connectivity, and range EEG (rEEG) features. Subsequently, they used feature selection techniques such as information gain, chi-square, analysis of variance, and Gini Index (GI) to rank the features by calculating the information entropy and deviation degree of each feature subset from the original data. The 600 most important features were then selected and fed into an SVM classifier, while Maniruzzaman [73] used the Least Absolute Shrinkage and Selection Operator (LASSO) to select final features and perform classification.

we extensively compared our method with the aforementioned relevant methods in terms of accuracy, sensitivity and specificity in ADHD classification. The specific results are presented in Table 3.

It can be observed that our method holds an advantage in nonlinear analysis. Additionally, in the case of basic time-frequency domain features, our employed nonlinear topological descriptor better captures the nonlinear dynamics of ADHD EEG, achieving effective classification.

However, as indicated in Table 2, our method shows a disadvantage when confronted with combined features, especially those using dimensionality reduction techniques and correlation-based comparisons. This suggests that individual nonlinear topological features alone may not fully capture the EEG state of ADHD patients. It underscores the necessity of combining time-frequency domain features with other nonlinear features for optimal performance. Nonetheless, this also underscores the feasibility of using topological feature descriptors for ADHD EEG classification, showcasing significant potential in EEG-related nonlinear analysis.

### *4.3. Clinical Assessment and Future Work*

Although after extensive experiments and tests, we have demonstrated that the topological features proposed in this paper have some effect in distinguishing ADHD EEG from normal human EEG, these results are based on mathematical extrapolations and dataset testing, and have not gained clinical practice. For this reason, we collaborated with the Department of Neurology, Shanxi Medical University, and invited some neuroscientists to comment on the results we obtained.

Most neuroscientists are concerned about feature interpretability versus clinical understanding. While EEG features can be proposed, however, the biological interpretability of these features may be more ambiguous to the clinician. Clinicians need to see not only the mathematical differences in EEG features during the diagnostic process, but also understand how these features correspond to the clinical manifestations of ADHD. While the algorithm is effective in capturing non-linear features in the EEG signal, it remains unclear how to correlate these topological features with core symptoms of ADHD, such as attention deficit and impulsivity.

Secondly, neurologists have suggested that simplifying channel selection or optimisation may be the next step towards improvement. In a clinical setting where time and resources for EEG analysis are limited, the use of EEG devices with 19 or more channels may be overly complex in some cases, and although the six channels of interest were selected by t-test in the paper, further research could explore ways to simplify the EEG channel setup while maintaining diagnostic accuracy, thereby improving the method's operability in real-world clinical settings.

The final problem is that more disease data is needed to improve the generalization performance of the model. In clinical practice, the complexity of patient symptoms and the uncertainty of the environment will greatly affect the accuracy of diagnosis. Therefore, the model should be continuously revised in large-scale practice. At the same time, combined with the first suggestion, that is, to find the biological significance of the feature, this method will have certain clinical use value.

In summary, our next plan intends to collect real disease data and modify the model accordingly, as well as to further collaborate with relevant neuroscientists to try to interpret the neurobiological significance of these topological features in the context of clinical knowledge, in order to make it easier for doctors to apply them to practical diagnosis.

## 5. Conclusion

In this paper, we utilized the framework of Topological Data Analysis to propose an improved method suitable for multi-channel ADHD EEG analysis. Following the reconstruction of the phase space to obtain signal point clouds, we employed k-PDTM to reconstruct the ideal point cloud structure. Additionally, in the remapped persistence diagrams, we utilized the MKDE method to filter outliers, ensuring the robustness of our approach. Furthermore, we discussed the influence of persistence points on classification results and conducted experiments using different weighting functions in the PI method. During validation, we tested and evaluated the effectiveness of this nonlinear topological descriptor using the IEEE ADHD dataset. The results demonstrated that compared to other nonlinear descriptors, utilizing topological features for ADHD classification yielded higher accuracy. However, our method showed a disadvantage when compared to feature combination approaches. In the future, our goal is to explore the integration of nonlinear topological features with other EEG characteristics for multichannel joint analysis. We aim to embed this algorithm into EEG acquisition devices, developing a system capable of rapid detection and identification of ADHD patients. We also plan to conduct clinical trials to validate this system. Furthermore, we intend to improve and extend this approach to other EEG tasks or the rapid recognition and detection of other neurological disorders.

Table 3. Comparison of classification performance of relevant features, including linear features, nonlinear features and their combinations

| Feature types | Method | Accuracy (%) | Specificity (%) | Sensitivity (%) |
|---|---|---|---|---|
| Non-linear features | ShanEn [59] | 76.4 | none | none |
| | SampEn [59] | 65.7 | none | none |
| | DispEn [59] | 68.6 | none | none |
| | MSE [59] | 67.1 | none | none |
| | CMSE | 71.90 | 70.00 | 73.77 |
| | Ours (topological feature) | 78.27 | 75.63 | 80.62 |
| Other features | Regional connectivity [71] | 53.0 | 62.0 | 49.0 |
| Combination of features | LLE + ApEn + Petrosian [58] | 78.60 | none | none |
| | ANOVA + Chi-square + Gini Index + Information Gain [72] | 76.86 | 76.88 | 76.86 |
| | Morphological Features + Time-domain Features [73] | 94.2 | 90.2 | 93.3 |

## Acknowledgements

The work was supported by the Basic Research General Program of Shanxi Province (No.202303021221186) , National Natural Science Foundation of China (No.62001430), and Shanxi Postgraduate Innovation Programme(No.2023SJ208).

## Author contributions

Tianming Cai proposed the idea, wrote the algorithm and wrote the article; Guoying Zhao helped to write the article; Junbin Zang assisted in the algorithm writing and data validation; Zongchen was responsible for the data collation, Zhidong Zhang and Chenyang Xue provided the experimental support and funding.

## Competing interests

The authors declare no competing interests.